\documentclass[letterpaper, 10 pt, conference]{ieeeconf} 

\IEEEoverridecommandlockouts 
\usepackage[dvipsnames]{xcolor} 
\usepackage{amsmath} 
\usepackage{amsfonts} 
\usepackage{graphicx}
\usepackage[font=small]{caption}
\usepackage{cuted} 
\usepackage{array} 
\usepackage{microtype}

\DeclareMathOperator*{\argmax}{argmax} 
\DeclareMathOperator*{\argmin}{argmin} 

\title{\LARGE \bf
Task-Aware Robotic Grasping by evaluating Quality Diversity Solutions through Foundation Models
}

\author{Aurel X. Appius$^{*1}$, Émiland Garrabé$^{1}$, Francois Hélénon$^{1}$, Mahdi Khoramshahi$^{1}$,\\ Mohamed Chetouani$^{1}$, and Stéphane Doncieux$^{1}$
\thanks{ $^{*}${Primary contributor}}
\thanks{ $^{1}${Sorbonne Université, CNRS, Institut des Systèmes Intelligents et de Robotique, ISIR, F-75005 Paris, France \tt \{appius, garrabe, helenon, khoramshahi, chetouani, doncieux\}@isir.upmc.fr} }
}%

\begin{document}
\maketitle
\thispagestyle{empty}
\pagestyle{empty}
\begin{strip}
  \centering
  \includegraphics[width=\textwidth]{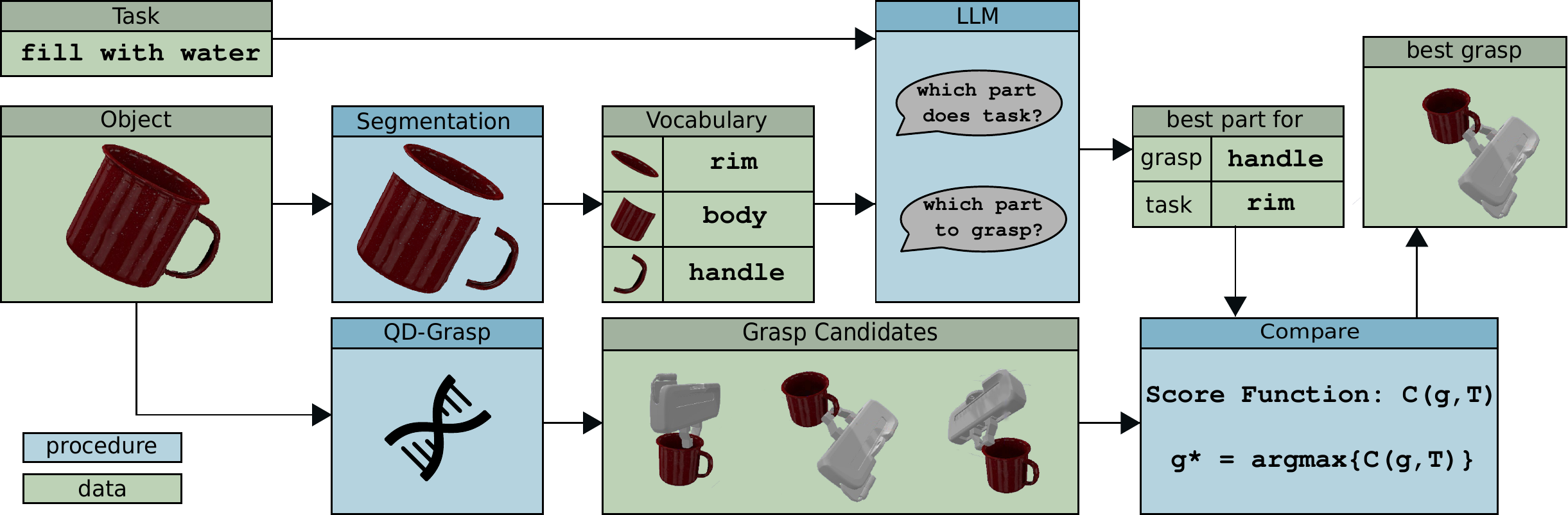} 
  \captionof{figure}{The proposed workflow for our task-aware grasping framework. An object and a task are given to the model, which are then segmented into a labeled dictionary of subparts. The relevant subparts for grasping and task execution are determined by prompting a Large Language Model (LLM). Finally, Quality Diversity algorithms generate grasp candidates for the object. The ideal grasp for a given task is found by maximizing a score function that rewards grasping on the suggested grasp subpart, maintaining a distance from the task subpart, while encouraging a high grasp force.}
  \label{fig:full}
\end{strip}


\begin{abstract}
Task-aware robotic grasping is a challenging problem that requires the integration of semantic understanding and geometric reasoning. This paper proposes a novel framework that leverages Large Language Models (LLMs) and Quality Diversity (QD) algorithms to enable zero-shot task-conditioned grasp synthesis. The framework segments objects into meaningful subparts and labels each subpart semantically, creating structured representations that can be used to prompt an LLM. By coupling semantic and geometric representations of an object's structure, the LLM's knowledge about tasks and which parts to grasp can be applied in the physical world. The QD-generated grasp archive provides a diverse set of grasps, allowing us to select the most suitable grasp based on the task. We evaluated the proposed method on a subset of the YCB dataset with a Franka Emika robot. A consolidated ground truth for task-specific grasp regions is established through a survey. Our work achieves a weighted intersection over union (IoU) of 73.6\% in predicting task-conditioned grasp regions in 65 task-object combinations. An end-to-end validation study on a smaller subset further confirms the effectiveness of our approach, with 88\% of responses favoring the task-aware grasp over the control group. A binomial test shows that participants significantly prefer the task-aware grasp.
\end{abstract}
\vfill\null

\begin{figure*}
  \centering
  \includegraphics[width=\textwidth]{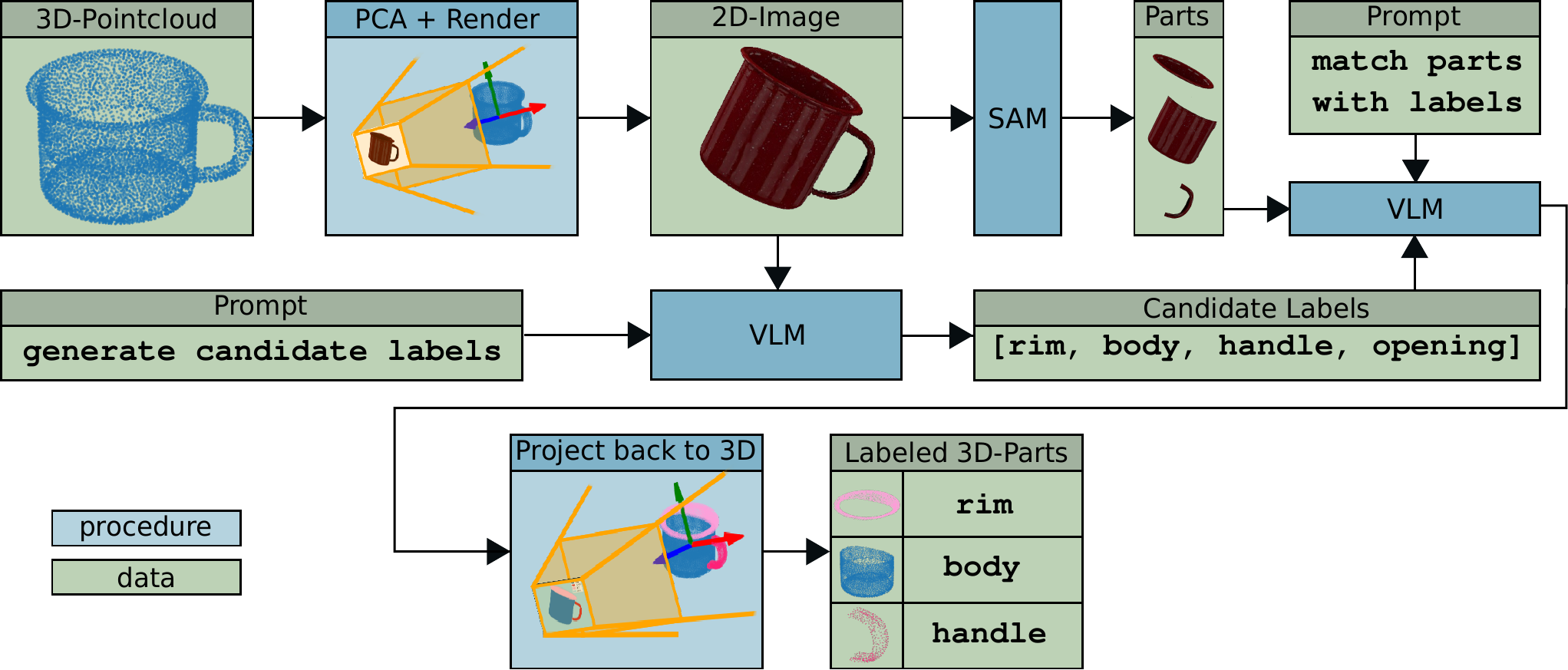} 
  \caption{The zero-shot semantic fine-grained segmentation is achieved by performing a Principal Component Analysis (PCA) and rendering the 3D Object along its most variance-explaining axes. The 2D Image is then segmented by {\tt segment-anything} \cite{kirillov2023segment} and labeled using {\tt GPT-4o} \cite{openai2024gpt4technicalreport} as a Vision Language Model (VLM). A projection back to 3D is used to generate the labeled semantic 3D subparts.}
  \label{fig:segmentation_pipeline}
\end{figure*}


\section{INTRODUCTION}
Robotic grasping, a fundamental aspect of manipulation, remains a complex and partially unsolved problem due to the wide range of object shapes, textures, and dynamic interaction requirements \cite{kroemer2020reviewrobotlearningmanipulation}. Beyond ensuring stability, effective grasping often requires consideration of the broader context, particularly the task that follows. The grasp chosen for an object is influenced not only by its geometric properties but also by its functional requirements and the semantic understanding of its purpose. Traditional grasp planning approaches often aim to identify stable or feasible grasps, largely independently from the object's broader context or its intended use \cite{Kleeberger2020}. This limitation creates a gap between robotic grasping capabilities and the precise demands of real-world applications, where the appropriateness of a grasp is inherently tied to the task that follows, such as pouring liquid from a cup, using a tool for assembly, or positioning an object for inspection. Addressing these dependencies is critical to improving task-oriented robotic manipulation. \\
\newpage


\section{RELATED WORK}
\subsection{Affordances}
The concept of affordances was introduced by Gibson in 1979 \cite{gibson1979ecological} and lays a theoretical foundation in robotics for bridging perception with action, allowing robots to understand the functional properties of objects beyond their geometric or visual characteristics. This theoretical basis inspired early task-oriented grasping approaches like \cite{fang2018learningtaskorientedgraspingtool} and \cite{chen2022learning6doftaskorientedgrasp} that can generate task-aware grasping poses using supervised learning methods. \cite{plonka2024learningspatialbimanualaction} model spatial affordance constraints for bimanual manipulation, reinforcing the role of structured affordance reasoning in grasping. Unfortunately, these methods require expensive annotated data and struggle with generalizing to unseen object categories.
\subsection{Large Language Models}
Large language models (LLMs) such as {\tt GPT-4o} \cite{openai2024gpt4technicalreport} have shown impressive abilities in tasks such as test-taking, computer vision, and natural language processing, which made them popular in both research and practical applications. LLMs excel in generalization and few-shot reasoning due to their vast encoded knowledge \cite{brown2020language}, making them highly suited for robotic applications where adaptability and contextual understanding are crucial \cite{firoozi2023foundationmodelsroboticsapplications}. Their ability to demonstrate a form of ``common sense'' enables them to interpret ambiguous scenarios and respond appropriately, offering significant potential for developing models that generalize far more effectively \cite{bommasani2022opportunitiesrisksfoundationmodels}. Despite these promising advantages, the use of this information in robotics is challenging due to the difficulty in grounding abstract task concepts in physical contexts \cite{liu2024aligningcyberspacephysical}. Given these opportunities and challenges, several works have demonstrated the potential of LLMs in grasping \cite{huang2024copageneralroboticmanipulation} and manipulation \cite{huang2024manipvqainjectingroboticaffordance}, underscoring their relevance and applicability for task-aware grasping.
\subsection{Task-Aware Grasping}
Given the limitations of supervised learning approaches and the promising qualities of LLMs, \cite{tong2024ovalpromptopenvocabularyaffordancelocalization} proposed to ground affordances using foundation models and then forward the gained insight into a grasp engine. Significant efforts by \cite{tang2023graspgptleveragingsemanticknowledge, tang2024foundationgraspgeneralizabletaskorientedgrasping} leveraged foundation models to identify both where and how to grasp objects. However, these methods still depend on manually annotated expensive datasets. \cite{mirjalili2023langraspusinglargelanguage} mitigate this by using an LLM to identify graspable object parts, but doesn't consider the subsequent task. \cite{cai2024visualimitationlearningtaskoriented} employ implicit neural fields for task-oriented grasping based on multiple human demonstrations.
\subsection{QD-Grasp}
QD-Grasp \cite{huber2023qualitydiversitysparsereward} is a novel approach to grasp synthesis that uses quality diversity algorithms to generate a diverse set of stable and high-quality grasps. By simultaneously optimizing for diversity and stability, QD-Grasp produces an extensive archive of distinct yet reliable grasp poses. This approach enables efficient sampling of a large number of grasps, facilitating the creation of comprehensive datasets suitable for policy learning \cite{huber2024qdgsetlargescalegrasping}. Considering that grasping is inherently linked to the subsequent task, the incorporation of task awareness into grasp synthesis is therefore essential to fully leverage the potential of QD-Grasp. Grasp archives should not only be varied and robust, but also contextually relevant.
\subsection{Perspective}
Our approach bridges the gap between high-level task understanding and robotic grasping by introducing a novel method to segment objects into subparts and give each subpart a semantic label. LLMs can identify the subpart to be grasped based on the task, which is linked to physical regions through the segmentation. By analyzing the QD generated grasp archive we can determine the optimal task-aware grasp pose. This direct alignment of LLM-encoded knowledge about grasping locations with the physical representation of the object is, to our knowledge, unprecedented in the literature. Our training-free approach provides a scalable and efficient way to generate task-aware grasps without requiring any annotated data.


\section{METHODOLOGY}
This section presents our methodology for leveraging LLMs to condition grasps on a given task. We propose a novel open-vocabulary 3D segmentation method that we use to identify task-relevant subparts by prompting an LLM.
\subsection{Problem Statement}
We define the grasp configuration space \( \mathcal{G} \) as the set of all possible 6 degree-of-freedom (DoF) grasps, where each grasp \( g = (p, R) \) is specified by position \( p \in \mathbb{R}^3 \) and orientation \( R \in SO(3) \). Let \( A \subset \mathcal{G} \) represent the grasp archive generated by the Quality Diversity algorithm \cite{huber2024speeding6dofgraspsampling}. Let \( C(g, \mathcal{T}) \) represent the task-compatibility score for grasp \( g \) and task \( \mathcal{T} \). We want to design \( C(g, \mathcal{T}) \) so that it captures the requirements of the tasks and ranks the grasps according to their compatibility with the task \(\mathcal{T}\). With the defined function \( C(g, \mathcal{T}) \), the optimal task-aware grasp \( g^* \) can be found through an optimization problem: 
\[
g^* = \argmax_{g \in A} C(g,\mathcal{T}),
\]
The goal is to define \( C(g, \mathcal{T}) \) in a way that ensures the optimization yields an optimal grasp \( g^* \).

\begin{figure}
  \centering
  \includegraphics[width=\columnwidth]{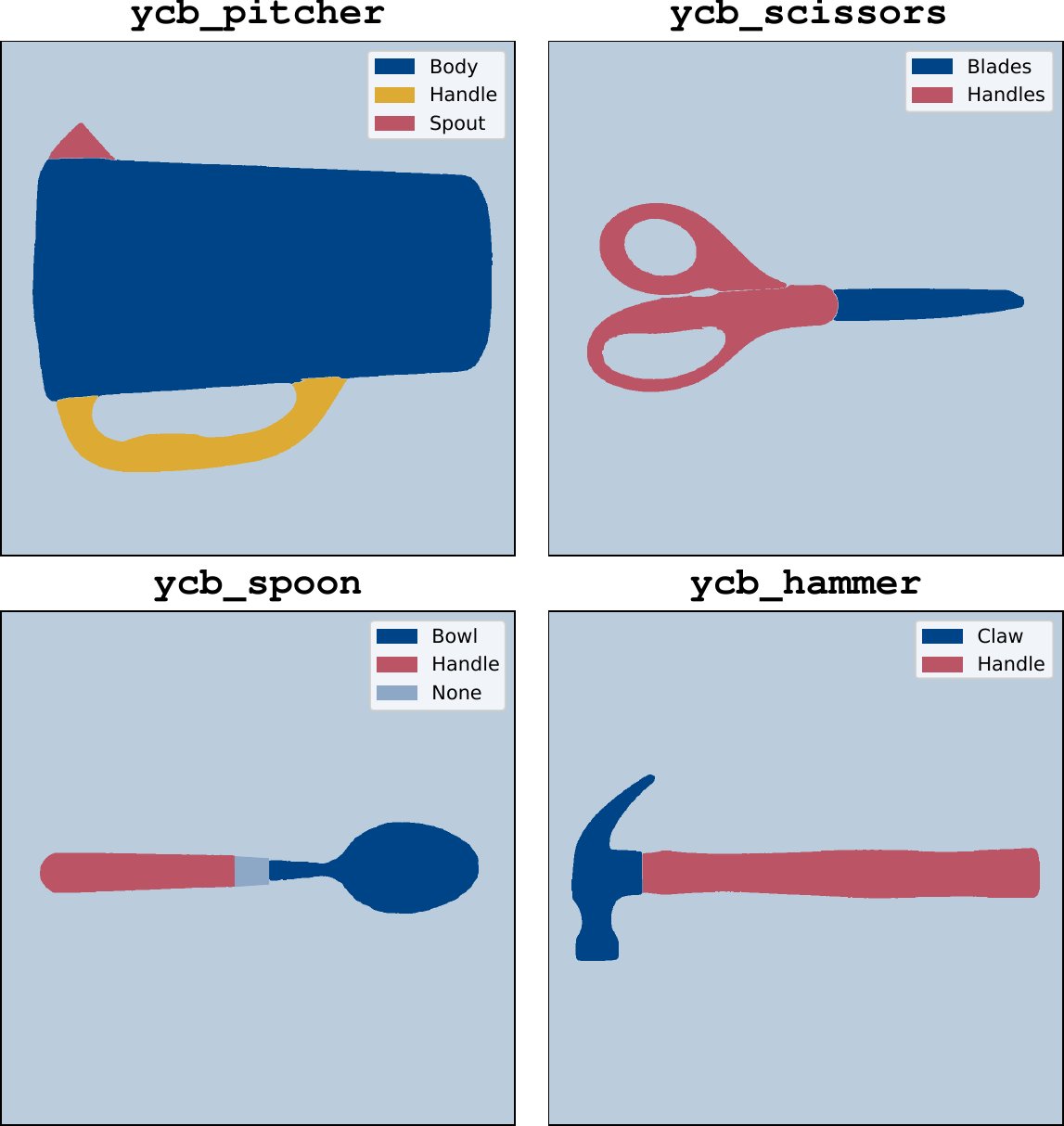} 
  \caption{Examples of labeled segmentation masks generated by the grasp segmentation pipeline using {\tt GPT-4o} \cite{openai2024gpt4technicalreport} as the VLM.}
  \label{fig:segmentations}
\end{figure}
\subsection{Segmentation} \label{Methodology:Segmentation}
An object \( O \) is defined by a combination of a semantic label \( L \) (for example, \textit{``mug''}) and its geometric properties represented by a point cloud \( \mathcal{P} \in \mathbb{R}^{N \times 3} \). Thus, an object is characterized as:
\[
O = (L, \mathcal{P}).
\]
For the object \( O \), our objective is to generate a set of subparts, denoted as the vocabulary \( V(O) \). Each subpart \( s_i \) within this vocabulary is characterized by a semantic label \( L_i \) and a point cloud \( \mathcal{P}_i \) that captures its shape and relative pose within the object. The vocabulary is defined as:
\begin{align*}
V(O) &= \{ s_i \mid s_i = (L_i, \mathcal{P}_i) \}, \\
\text{where} \quad & \bigcup_i \mathcal{P}_i = \mathcal{P} \quad \text{and} \quad \mathcal{P}_i \cap \mathcal{P}_j = \emptyset, \; i \neq j.
\end{align*}
The vocabulary \( V(O) \) is generated by performing a semantic segmentation on the object \( O \). As open-vocabulary semantic segmentation on 3D parts is unsolved and highly challenging \cite{xiao20243dopenvocabularypanopticsegmentation}, we propose the following method.
\begin{enumerate}
  \item \textbf{Render and Segment the Point Cloud:} Render the 3D point cloud to generate image \( \mathcal{I} = \text{render}(P) \). This is done by performing a Principal Component Analysis (PCA) on the point cloud to determine the object viewing angle: the two dimensions with the highest variance are selected for rendering, while the camera is aligned along the dimension with the least variance. Furthermore, the rendered 2D image is segmented using {\tt segment-anything} \cite{kirillov2023segment} to obtain 2D masks.
  \item \textbf{Generate Candidate Labels:} Use a Vision-Language Model (VLM) such as {\tt GPT-4o} \cite{openai2024gpt4technicalreport} to generate a set of candidate semantic labels \( \{ L_1, L_2, \ldots, L_n \} \) for the object \( O \), based on the overall label of the object \( L \) and the image \( \mathcal{I} \). As we intend to find graspable and task-relevant subparts, we specify in the VLM prompt that the candidate labels should be graspable or task-relevant subparts of the object \( O \).
  \item \textbf{Assign Labels to 2D Segments:} Use a VLM to assign a semantic label \( L_i \) to each segmented 2D region from the set of candidate labels \( \{ L_1, L_2, \ldots, L_n \} \).
  \item \textbf{Project Labeled Segments into 3D:} Combine duplicate and remove overlapping regions, then project the labeled 2D segments back into 3D to obtain the labeled subparts \(s_i = (L_i, \mathcal{P}_i) \) and, thus, the vocabulary \(V(O)\).
\end{enumerate}
This approach leverages 2D segmentation with {\tt segment-anything}\cite{kirillov2023segment} and combines it with VLMs to efficiently identify and label the components of an object in 3D. The segmentation pipeline is illustrated in figure \ref{fig:segmentation_pipeline}.
\subsection{Task Conditioning} \label{subsec:task_condition}
We construct a prompt \( P \) for the task \( \mathcal{T} \) and the set of subpart labels \( \{ L_1, L_2, \ldots, L_n \} \). The prompt \(P\) asks the LLM to return the following two labels:
\begin{itemize}
\item \( L_\text{grasp} \): The label of the subpart best suited for grasping.
\item \( L_\text{task} \): The label of the subpart responsible for performing the task.
\end{itemize}
\[
(L_\text{grasp}, L_\text{task}) = \text{LLM}(P),
\]
The provided labels \( L_\text{grasp} \) and \( L_\text{task} \) are used to construct the score function \( C(g,\mathcal{T}) \). We intend to minimize the chance of the gripper interfering with the task by rewarding grasps with a large distance between the grasp points and the task subpart \(L_\text{task}\). We therefore define \(d_\text{task}\) as the minimal euclidean distance between the gripper-object contact point \( p_g \) and the subpart point cloud \(\mathcal{P}_{L_\text{task}} \).
\[
d_{\text{task}} = \min_{p \in \mathcal{P}_{L_{\text{task}}}} \| p_g - p \|_2
\]
Additionally, we determine the grasped subpart \(L_{g}\) of grasp \(g\) by finding the label of the point cloud closest to the grasp point \(p_g\):
\[
L_g = L_i \quad \text{where} \quad i = \argmin_{j \in \{ j' \mid (L_{j'}, \mathcal{P}_{j'}) \in V(O) \}} \| p_g - \mathcal{P}_j \|_2
\]
Finally, we define the score function \( C(g,\mathcal{T}) \) as:
\[
\fbox{
    $
    C(g,\mathcal{T}) = \begin{cases}
        K_\text{force} \cdot F + K_\text{dist} \cdot d_\text{task},& \text{if }L_g = L_\text{grasp} \\
        0,              & \text{otherwise}
    \end{cases}
    $
}
\]
where \(F\) is the contact force, that we obtain from the pybullet simulation engine used to generate the QD grasp archive. \(K_\text{force}\) and \(K_\text{dist}\) are gains that balance the influence of the contact force and the distance from \(L_\text{task}\). This formulation defines \( C(g,\mathcal{T}) \) as a score function that, when maximized, discards all grasps not on \( L_\text{grasp} \) while encouraging strong grasps with distance to \( L_\text{task} \), therefore giving us the task-conditioned grasp \(g^*\).
\[
g^* = \argmax_{g \in A} C(g,\mathcal{T})
\]
Figure \ref{fig:full} provides a visual overview of the described pipeline. 


\section{EXPERIMENTS}
\begin{figure}
  \centering
  \includegraphics[width=\columnwidth]{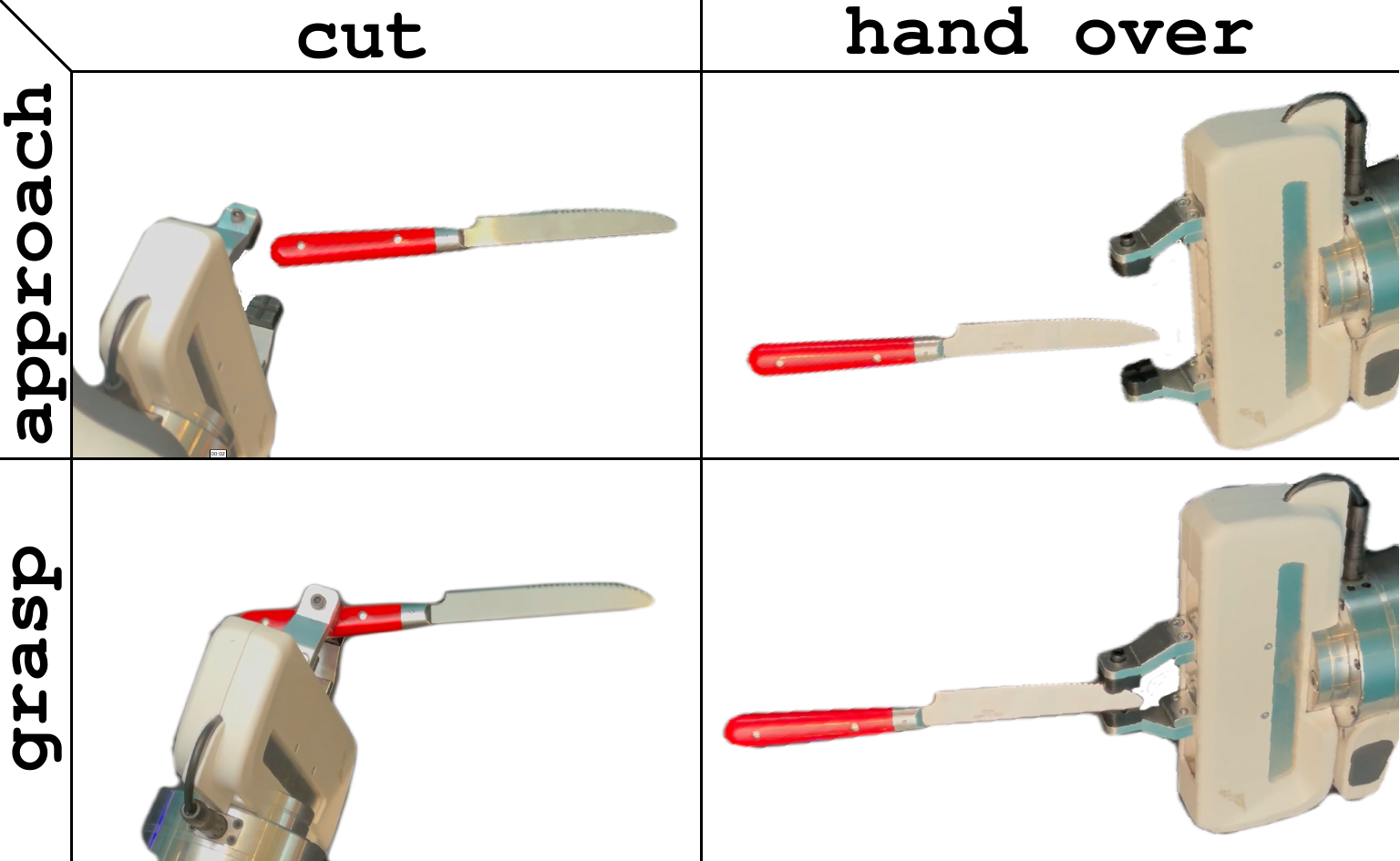} 
  \caption{The Franka Emika robotic arm, equipped with a Panda 2-DoF gripper, used for conducting experimental evaluations of task-specific grasping. The figure illustrates the progression of two distinct grasps, each conditioned on a separate task. The left grasp is for the task of cutting something with the knife, and the right one is for handing the knife over to someone.}
  \label{fig:franka}
\end{figure}
\subsection{Experimental Setup}
To experimentally validate the task-aware grasping framework, the method was deployed on a Franka Emika arm with a Panda 2-DoF gripper as seen in Figure \ref{fig:franka}. The system asks the user to specify the task \(\mathcal{T}\). The presented framework determines the task part \( L_\text{task} \) and grasp part \( L_\text{grasp} \) which are used to create the score function \( C(g,\mathcal{T}) \) as described in section \ref{subsec:task_condition}. The precomputed grasp archive \(A\) is loaded and evaluated, resulting in an optimal grasp \( g^* \). \\
Through \(\tt{Moveit!}\) \cite{coleman2014reducingbarrierentrycomplex} motion planning, the approach trajectory is computed and the grasp is executed. We chose {\tt GPT-4o} \cite{openai2024gpt4technicalreport} as both the VLM in the segmentation pipeline and the LLM for the grasp region suggestions. All the following results have been produced with this set-up.
\subsection{Object Dataset}
The method was evaluated using a subset of the {\tt YCB dataset} \cite{ycb_pub}, selecting objects that encompass a wide range of geometries and visual appearances. Particular preference was given to objects with multiple affordances, especially those where the grasp location would vary depending on the intended task. This selection enables us to test the framework's ability to adapt to task-specific grasping requirements. We defined a large set \(S_\text{large}\) of 65 task-object combinations for the validation of the grasp region prediction, described in section \ref{results:grasp_region_prediction}, as well as a small set \(S_\text{small}\) of 7 task-object combinations for end-to-end validation on the real robot, as described in section \ref{results:end_to_end_validation}. Table \ref{tab:dataset} shows the combination of tasks and objects, where objects and tasks from \(S_\text{small}\) are colored in red.

\begin{table}
  \centering
  \begin{tabular}{|m{0.12\textwidth}|m{0.3\textwidth}|} \hline
    \textbf{Name} & \textbf{Tasks} \\ \hline
    \texttt{\textcolor{red}{screwdriver}} & hand over, place, insert, stir, shake, \textcolor{red}{screw}, prick \\ \hline
    \texttt{powerdrill} & hand over, place, shake, screw \\ \hline
    \texttt{bowl} & hand over, place, pour, scoop, fill \\ \hline
    \texttt{\textcolor{red}{pitcher}} & hand over, place, pour, shake, \textcolor{red}{fill} \\ \hline
    \texttt{\textcolor{red}{spatula}} & hand over, place, hang, stir, shake, \textcolor{red}{flip} \\ \hline
    \texttt{fork} & hand over, place, cut, prick \\ \hline
    \texttt{spoon} & hand over, place, stir, scoop, fill \\ \hline
    \texttt{\textcolor{red}{scissors}} & hand over, place, cut, insert, \textcolor{red}{hang} \\ \hline
    \texttt{hammer} & hand over, place, press, hammer \\ \hline
    \texttt{marker} & hand over, place, insert, squeeze, write \\ \hline
    \texttt{\textcolor{red}{mug}} & hand over, place, hang, pour, shake, scoop, \textcolor{red}{fill} \\ \hline
    \texttt{\textcolor{red}{knife}} & \textcolor{red}{hand over}, place, \textcolor{red}{cut}, stir, shake, prick \\ \hline
  \end{tabular}
  \caption{A comprehensive list of the task-object combinations of \(S_\text{large}\). The subset \(S_\text{small}\) is indicated in \textcolor{red}{red}.}
  \label{tab:dataset}
\end{table}


\section{RESULTS}
\subsection{Grasp Object Segmentation}
The segmentation pipeline demonstrates promising results by effectively distinguishing fine-grained subparts of objects in open-vocabulary scenarios. However, evaluating the accuracy of these segmentation masks is a challenging task due to the open nature of both the objects and the vocabulary used. Unlike fixed-category segmentation problems, our method is designed to handle arbitrary object classes and corresponding labels, making quantitative evaluation difficult. However, qualitative assessments suggest that the proposed pipeline generates segmentation masks that are useful for task-aware grasping, as shown in Figure \ref{fig:segmentations}.
\subsection{Grasp Region Prediction} \label{results:grasp_region_prediction}
\begin{figure}
  \centering
  \includegraphics[width=\columnwidth]{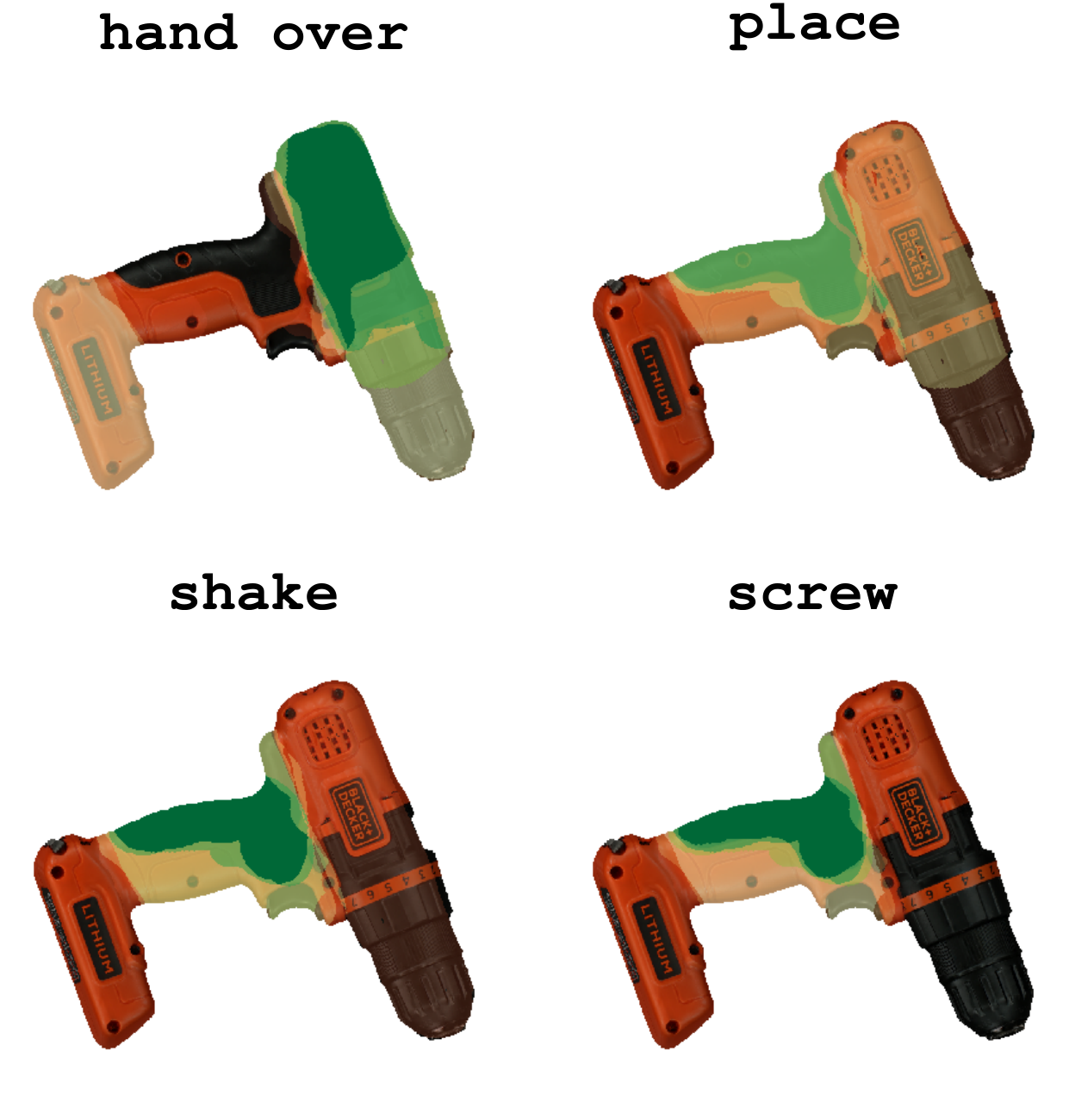} 
  \caption{Consolidated ground truth of grasping regions that were determined in the survey. For each task, the participants had to highlight a region were they would grasp the object in order to do a subsequent task. The figure shows an aggregation of the data that was collected. The intensity of green indicates the regions popularity for the specific task.}
  \label{fig:ground_truth}
\end{figure}

Establishing a ground truth for task-conditioned grasping is challenging due to the subjective nature of task-specific requirements. Unlike conventional grasp planning metrics, which focus on stability or feasibility, task-aware grasping lacks a universally accepted evaluation method. While task performance could serve as an alternative metric, directly evaluating grasps through real-world task execution is often impractical. Given this constraint, we use human intuition as an alternative reference. To capture human preferences, we conducted a survey on where to grasp an object based on the intended task. For every task-object combination in \(S_\text{large}\), we asked six participants to indicate their preferred grasp region, which we then combined into a consolidated ground truth. Figure \ref{fig:ground_truth} shows the ground truth for a power drill for different subsequent tasks.\\

When inferring the model for a certain task \(\mathcal{T}\), the label \(L_{\text{grasp}}\) is generated, which is coupled to a geometric subpart \(\mathcal{P}_{\text{grasp}}\) on the object through the vocabulary \(V(O)\). We compare this predicted grasp region with the consolidated ground truth by performing a weighted IoU. Weights are allocated based on the extent of overlap in participants' survey responses, where increased overlap signifies greater significance in the consolidated ground truth. Our model reached a \textbf{weighted IoU of 73.6 \%} with a precision of 91.1 \% and a recall of 73.6 \% in the data set \(S_\text{large}\). To ensure the robustness of these metrics and account for potential variations across runs, we repeated the experiment ten times and report the average performance across these runs.

\subsection{End-to-end Validation} \label{results:end_to_end_validation}
\begin{figure}
  \centering
  \includegraphics[width=\columnwidth]{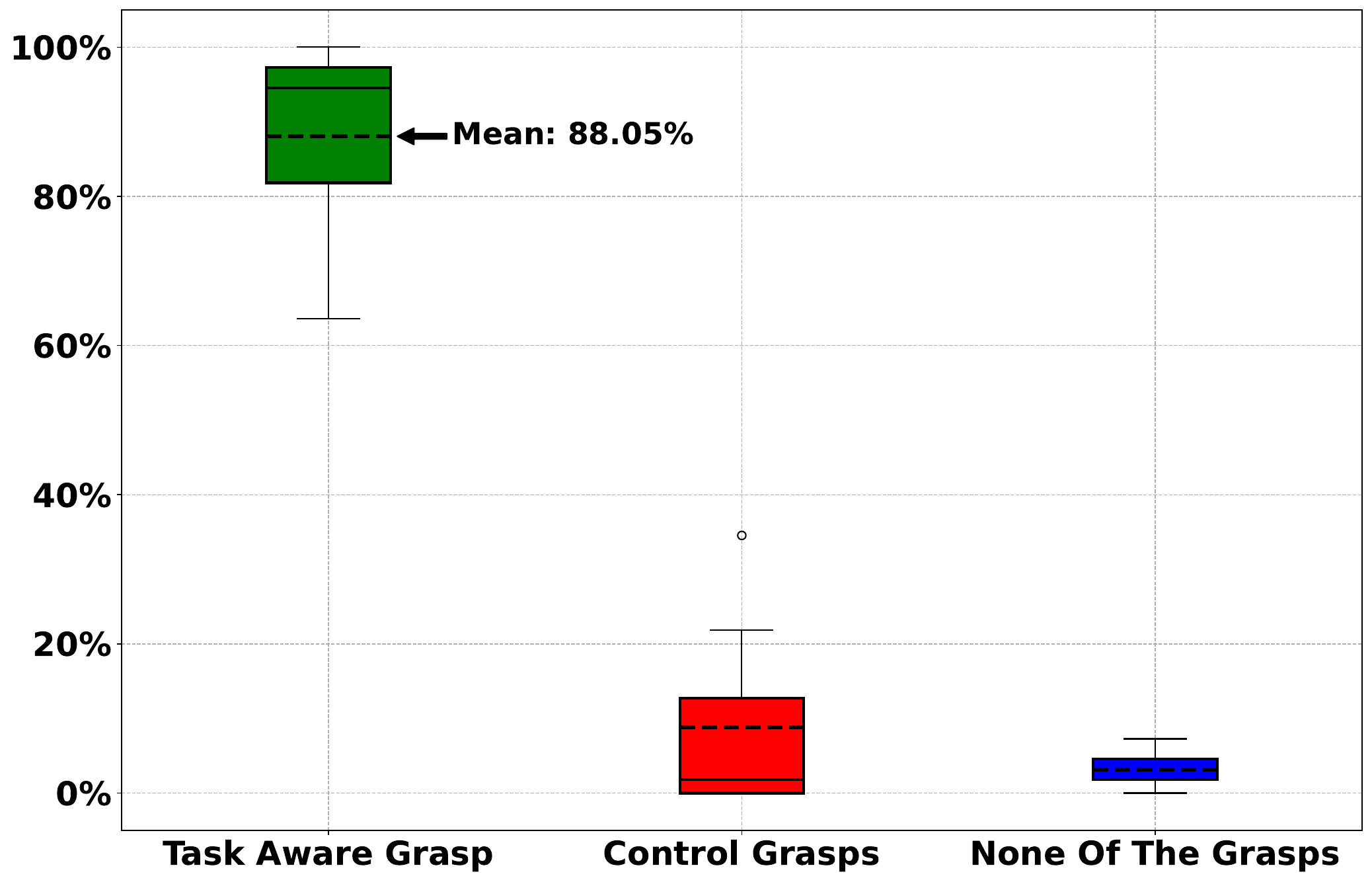} 
  \caption{Box plot showing participant selection percentages of the survey. The results show strong evidence that the task-awareness algorithm predicts grasps that are aligned with human intuition. }
  \label{fig:boxplot}
\end{figure}

A complete end-to-end validation was performed on \(S_\text{small}\), evaluating the full pipeline from task-object combination as input to task-conditioned grasp selection and human preference assessment as output. Similarly to the validation in section \ref{results:grasp_region_prediction}, the end-to-end validation also relies on human intuition, where no objective ground truth dataset is available. To validate the effectiveness of our approach, we determined the optimal task-aware grasp \(g^*\) according to the proposed approach and for each task-object combination
\[
g^* = \argmax_{g \in A} C(g,\mathcal{T}), \quad \text{with } K_\text{force} = 10, K_\text{dist} = 1
\]
and sampled three additional grasps \( g_1, g_2, g_3\) for the control group.
\[
\{ g_1, g_2, g_3 \} \text{ such that } C(g_i,\mathcal{T}) = 0, \quad \forall i \in \{1,2,3\}.
\]
We filmed these grasps and asked a group of participants to select their preferred grasp given the task \(\mathcal{T}\). They could choose from \(g^*, g_1, g_2, g_3\), or \(\tt{none}\). We hypothesize that they will align with the algorithm by favoring the task-aware grasp \(g^*\) over the alternatives. A total of 55 people participated in the survey, with \textbf{88\%} preferring the task-aware grasp, as shown in Figure \ref{fig:boxplot}. A binomial test, based on the null hypothesis that participants choose each grasp equally often, confirms that \textbf{the task-aware grasp is significantly preferred} (\(\mathbf{p < 0.001}\)).


\section{DISCUSSION \& LIMITATIONS}
Our approach introduces a new method for task-aware grasping by conditioning 6-DoF grasps on task requirements. By using a custom segmentation method, we can align the LLMs task knowledge with the grasping capabilities of the robot. However, several limitations persist, presenting opportunities for further research.
\subsection{Segmentation}
The PCA-based projection in the segmentation pipeline is a key limitation, making it difficult to segment concave or complex-shaped objects. For such geometries, the 2D projection may obscure details or merge distinct subparts, leading to inaccurate segmentation. Since the grasping of such objects itself remains a challenging and partially unsolved problem, this limitation is not overly restrictive. Advances in computer vision are ongoing, and even more effective segmentation methods will emerge in the future. Nonetheless, the segmentation pipeline should be evaluated against an established segmentation dataset to assess its effectiveness with more rigour.
\subsection{Further LLM interaction}
While considering the subparts of the objects for grasping is a strong first step, there is significant potential for improvement in the LLM-based grasp specification. For instance, the LLM could be prompted to determine whether a task requires a high or low grasp force, or even to reason about the optimal grasp orientation. Grounding orientation and incorporating it into the LLM’s reasoning could further enhance task-specific grasp selection. All of these ideas could be seamlessly integrated into the task-conditioned score function \( C(g,\mathcal{T}) \), leading to a more adaptive framework that better captures task-specific grasp requirements.

\section{CONCLUSION}
This work presents a zero-shot, task-aware robotic grasping framework that integrates semantic understanding with geometric reasoning, enabling robots to select grasps aligned with task requirements without any additional training. Our grasp region prediction achieved strong performance, with \textbf{73.6\%} IoU compared to a survey-based ground truth. Furthermore, in an end-to-end setting, a binomial test with \(\mathbf{p < 0.001}\) showed that \textbf{participants significantly preferred the task-aware grasp}. Future research should focus on refining segmentation for complex geometries, enhancing LLM interaction, and scaling to large datasets, as addressing these challenges will further strengthen the capabilities of task-aware grasping. Ultimately, this work takes an important step toward solving task-aware grasping, a fundamental aspect of robotic manipulation.


\section*{ACKNOWLEDGMENT}
This work was supported by the French Agence Nationale de la Recherche (ANR) (ANR-21-FAI1-0004) (Learn2Grasp) and the European Union Horizon Europe Framework Program, through the PILLAR (grant agreement 101070381) and euROBIN (grant agreement 101070596) projects.

\bibliography{IEEEabrv.bib,references.bib}{}
\bibliographystyle{IEEEtran} 
\end{document}